\documentclass[runningheads]{llncs}
\usepackage{graphicx}

\usepackage{tikz}
\usepackage{comment}
\usepackage{amsmath,amssymb} 
\usepackage{color}

\usepackage{booktabs}
\usepackage{multirow}
\usepackage{bm}
\newcommand{\etal}{\textit{et al.}}

\usepackage[symbol]{footmisc}

\usepackage[accsupp]{axessibility}  

\begin{document}

\pagestyle{headings}
\mainmatter
\def\ECCVSubNumber{3154}  

\title{LaT: Latent Translation with Cycle-Consistency for Video-Text Retrieval} 


\titlerunning{LaT: Latent Translation with Cycle-Consistency for Video-Text Retrieval}

\author{Jinbin Bai$^*$, Chunhui Liu$^{\dagger}$, Feiyue Ni$^*$, \\Haofan Wang, Mengying Hu, Xiaofeng Guo, Lele Cheng}

\authorrunning{J. Bai et al.}

\institute{MMU, Kuaishou Technology \\ \email{\{baijinbin,liuchunhui,nifeiyue,\\wanghaofan,humengyin,guoxiaofeng,chenglele\}@kuaishou.com}}

\footnotetext[1]{Interns at MMU, Kuaishou Technology.}
\footnotetext[2]{Corresponding author.}

\maketitle


\begin{abstract}

    Video-text retrieval is a class of cross-modal representation learning problems, where the goal is to select the video which corresponds to the text query between a given text query and a pool of candidate videos. The contrastive paradigm of vision-language pretraining has shown promising success with large-scale datasets and unified transformer architecture, and demonstrated the power of a joint latent space. Despite this, the intrinsic divergence between the visual domain and textual domain still remains, and projecting different modalities into a joint latent space may result in the distortion of information inside the single modality. To address this issue, we introduce a novel mechanism for learning the translation relationship between source modality space $\mathcal{S}$ and target modality space $\mathcal{T}$ without the need for a joint latent space, which bridges the gap between visual and textual domains. Additionally, to maintain cycle consistency between the translations, we adopt a cycle loss involving both forward translations from $\mathcal{S}$ to the predicted target space $\mathcal{T'}$, and backward translations from $\mathcal{T'}$ back to $\mathcal{S}$. Extensive experiments conducted on MSR-VTT, MSVD, and DiDeMo datasets demonstrate the superiority and effectiveness of our LaT approach compared with vanilla state-of-the-art methods.  
    
\end{abstract}

\begin{sloppypar}
\section{Introduction}

Video-text retrieval requires a bidirectional understanding between video and language, aiming to select the video which corresponds to the text query between a given text query and a pool of candidate videos, and vice versa. Cross-modal representation learning has witnessed an explosion of architectures to solve this task, be it single-stream~\cite{chen2020uniter} or dual-stream architectures~\cite{radford2021learning}, early fusion or late fusion. And the contrastive paradigm of vision-language pretraining~\cite{radford2021learning} has shown promising success with large-scale datasets and unified transformer architectures, which simply and crudely maps embeddings from different modalities to a common latent space via linear layers, thereby demonstrating the power of a joint latent space. 

\begin{figure}[t]
    \centering
    \includegraphics[width=.95\textwidth]{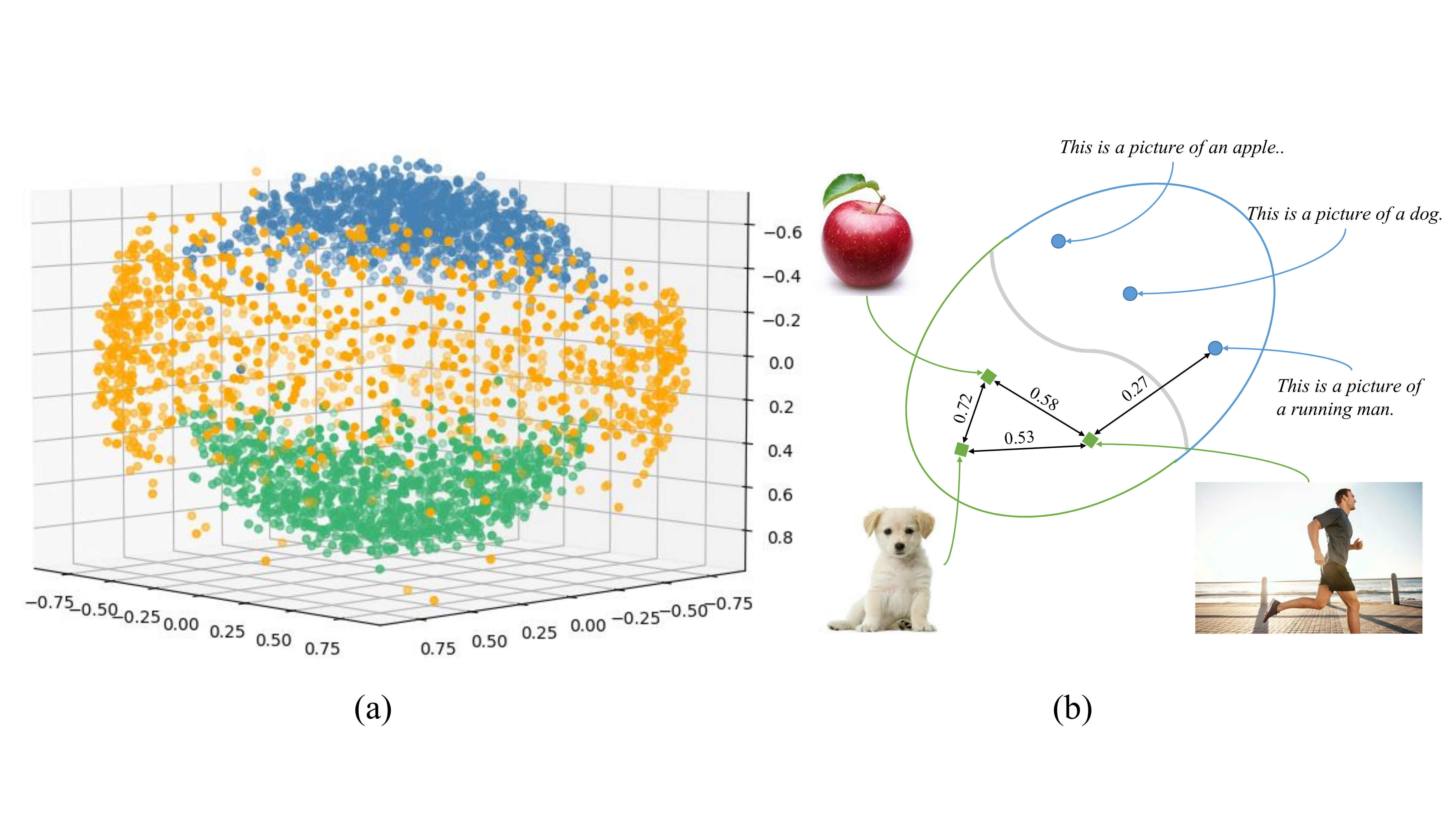}
    \caption{Figure (a) is a visualization of embeddings from different latent spaces, with green points and blue points represent latent embedding from CLIP visual encoder and textual encoder, respectively. And the orange points represent latent textual embedding from BERT. Figure (b) demonstrates the intrinsic gap between visual domain and textual domain by measuring the \textbf{cosine similarity} (where larger values indicate greater similarity) between pictures and their text descriptions. As an example, the similarity between picture \textit{Running man} and picture \textit{Apple} or picture \textit{Dog} is 0.58, 0.53 respectively, far exceeding the similarity between picture \textit{Running man} and text \textit{This is a picture of a running man.}, which is 0.27. This reveals a great divergence between the different modalities.}
    \label{Visualization}
\end{figure}

Despite this, the intrinsic gap between the visual domain and textual domain still remains, and projecting different modalities into a joint latent space may result in the distortion of intra-modal information (information inside the single modality). To illustrate our conjecture more intuitively, we randomly selected 1,000 image-text pairs from Google Conceptual Captions 3M dataset~\cite{sharma2018conceptual} and obtain their embeddings. Fig. \ref{Visualization}-(a) shows some visualization after dimensionality reduction with multidimensional scaling~\cite{borg2005modern}. The green points and blue points represent latent embedding from CLIP~\cite{radford2021learning} visual encoder and textual encoder, respectively, while the orange points represent textual latent embedding from BERT. The divergence between the distribution of the blue and orange points indicates that the distribution obtained by the textual representation model has changed during the feature alignment process, which confirms the conjecture of distortion of latent space. Additionally, the large gap between the distribution of the blue and green points suggests that in the process of supervising the formation of the visual model through the natural language model, CLIP is still unable to perfectly merge the latent spaces of two modalities with a simple linear projection. Fig. \ref{Visualization}-(b) provides further quantitative details about this intrinsic gap between the visual domain and textual domain. 

One potential solution to the existing divergence is to establish bridges connecting them. This can be done through techniques such as machine translation for different languages, or image-to-image translation using CycleGAN~\cite{zhu2017unpaired}. Our method aims to maximize the preservation of each original modality's unique qualities while minimizing the cost of unifying them.

More specifically, we refer to these bridges as a form of more general latent translation, which is natural and applicable in cross-modal learning. The query embeddings $Q_G$ and $Q_F$ are learnable parameters used to guide the translation processes $G: \mathcal{S} \xrightarrow{} \mathcal{T}$ from source modality $\mathcal{S}$ to target modality $\mathcal{T}$, and $F: \mathcal{T} \xrightarrow{} \mathcal{S}$ from $\mathcal{T}$ to $\mathcal{S}$. The magnitude of $Q_G$ and $Q_F$ are determined by the dimension of $T$ and $S$, respectively. Under the guidance of these $Q_G$ and $Q_F$, latent embeddings from domain $\mathcal{T}$ can be readily translated to domain $\mathcal{V}$, and vice versa. Consequently, for each paired embedding $t \in \mathcal{T}$ and $v \in \mathcal{V}$, we impose constraints $G(t) \approx v$ and $F(v) \approx t$ between the translated embedding and the desired target embedding. Moreover, after the cycle translation, the final target domain should be consistent with its original domain, whatever the intermediate domain is. To keep this cycle consistency, we add further constraints $v = G(F(v))$ and $t = F(G(t))$ between the source embedding and the final target domain which has been translated twice.

\begin{figure}[t]
    \centering
    \includegraphics[width=1\textwidth]{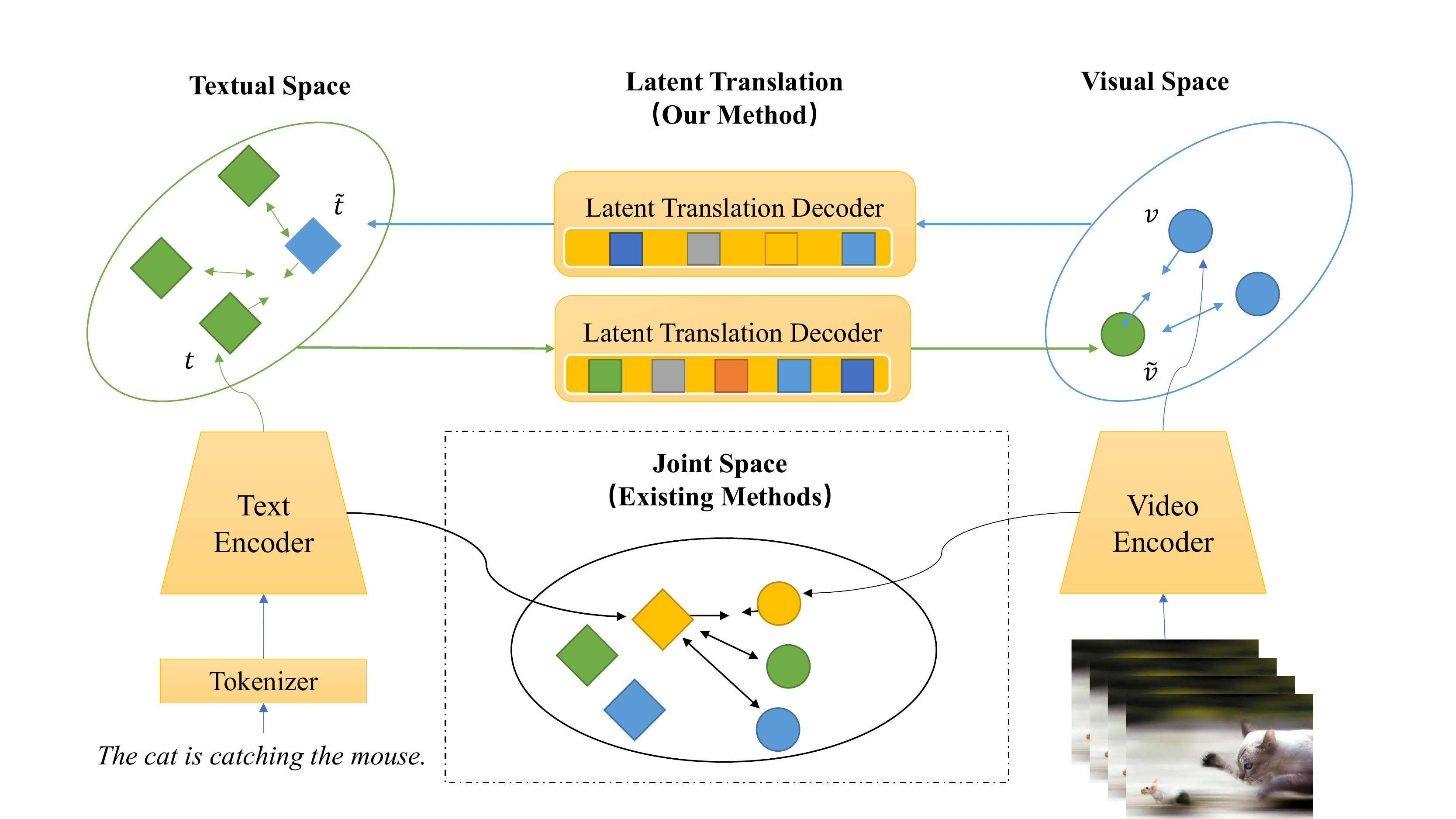}

    \caption{Overview of our latent translation method. The decoder above can transform a video $v$ to a meaningful embedding $\widetilde{t}$ in the textual space, and the decoder below can transform a sentence $t$ to a meaningful embedding $\widetilde{v}$ in the visual space. }
    \label{LaT}

\end{figure}

In this paper, we take a step towards building bridges between visual and textual modalities, by proposing a dual decoder architecture which utilizes the learnt queries~\cite{carion2020end} to translate between the two domains (Fig. \ref{LaT}). This allows us to convert an embedding from textual space to one within visual space, and then back again, and we call this latent translation process cycle-consistent.

In summary, our main contributions can be summarized as follows:
\begin{itemize}
    \item[$\bullet$] We analyzed the intrinsic modality gap between visual and textual modalities through qualitative and quantitative methods.

    \item[$\bullet$] We presented a novel latent translation framework and incorporated constraints such as cycle-consistent loss and intra-modal contrastive loss to promote cross-modal translation.
    
    \item[$\bullet$] Through extensive experiments conducted on MSR-VTT, MSVD, and DiDeMo datasets, we demonstrate the superiority and effectiveness of our LaT approach compared with vanilla state-of-the-art methods for video-text retrieval.
\end{itemize}

\section{Related Work}

Our work builds on prior works in several domains, including dual stream architecture for cross-modal retrieval, video representation based on image encoder, and cycle-consistent learning methods.

\subsection{Cross-modal Retrieval}
Existing methods for cross-modal learning can be roughly categorized as single stream~\cite{li2020unicoder,li2019visualbert,li2020oscar,chen2020uniter,lu2019vilbert,sun2019videobert,lin2021vx2text,hu2021unit} and dual stream architectures~\cite{radford2021learning,jia2021scaling,liu2021hit,wang2021efficientclip}. Single stream architecture directly fuses visual and textual representations with a multi-modal transformer or independently extract visual and textual features then introduces a cross-modal attention to merge multi-modal information. Dual stream architecture employ two encoders to extract different modal representations and projects them into a common latent space with a similarity-based ranking loss, has become a recent trend for cross-modal retrieval. One illustrious work with dual stream architecture is CLIP~\cite{radford2021learning}. 

Dual stream models have a merit of efficient inference for downstream tasks such as image-text retrieval, since they can decouple and offline store pre-computed image/text features from encoders~\cite{gu2022wukong}. Assuming that we have M videos and N texts, when it comes to large-scale cross-modal retrieval tasks, single stream method usually requires $\Theta(MN)$ time complexity of intra-modal information exchange while dual stream method requires only $\Theta(M+N)$ time complexity. This makes real-time cross-modal retrieval practical with dual stream models.

\subsection{Video Encoder Backbone}
The early research on video representation learning utilize 2D convolutions on spatial features ~\cite{karpathy2014large} with a temporal sampling method \cite{wang2016temporal} to capture temporal information. Moreover, 3D convolutions are further proposed \cite{tran2015learning,ji20123d,liu2021selective} to jointly learn spatio-temporal information. The convergence of 3D kernels is further improved by initializing them from inflated 2D kernels \cite{carreira2017quo} (known as I3D models) and decoupling spatial and temporal convolutions in each block~\cite{tran2018closer} (known as R(2+1)D models). The success of the vision transformer on images also inspire the progress of the video encoder. \cite{arnab2021vivit,bertasius2021space} propose pure transformer architectures to model spatial-temporal feature. 

Another benefit of video transformers is that the main attention function that works on a group of local patches doesn't rely on inductive bias and makes it easy to be extended on a joint image and video input. The effective transformer proposed by Bain~\cite{bain2021frozen}, which is based on TimeSformer\cite{bertasius2021space}, can gracefully handle inputs of different length of videos and images (by treating images as a single-frame video).

\subsection{Cycle-Consistent Learning}
Inspired by machine translation in various languages, Zhu~\etal ~\cite{zhu2017unpaired} exploited unpaired image-to-image translations between two different visual domains with a cycle-consistency loss and an adversarial loss~\cite{goodfellow2014generative}. Several works have also transferred the cycle-consistency loss to image-text retrieval tasks~\cite{liu2019cyclematch,wu2018cycle,cornia2018towards}. CycleMacth~\cite{liu2019cyclematch} utilizes fully-connected layers to maintain inter-modal correlations and intra-modal consistency. DGH~\cite{wu2018cycle} introduces a deep generative approach for cross-modal retrieval to learn hash functions. The concept Cornia~\etal~\cite{cornia2018towards} proposed is pretty similar to us. However, their translation frameworks are based on convolution neural networks without any satisfactory ablation experiments or attempt to incorporate other architectures. 

Here we propose a brand new latent translation framework with cycle-consistency, which introduces the attention~\cite{vaswani2017attention} architecture and learnable queries~\cite{carion2020end}, for cross-modal retrieval.

\section{Method}
In this section, we will define our tasks (Section \ref{pd}), present architecture of our Latent Translation (LaT) (Section \ref{Model Architecture}), provide information on our latent translation decoder (Section \ref{decoder}) and elaborate on the supervised methods we designed (Section \ref{Loss}).

\subsection{Problem Definition} \label{pd}

Let a vector $\Bar{v}$ denote a raw video and a vector $\Bar{t}$ denote a raw text. We use a vision encoder $E_v$ and a language encoder $E_l$ to get the latent feature $v \in \mathcal{V}$ and $t \in \mathcal{T}$, where $\mathcal{V}$ and $\mathcal{T}$ are visual and textual latent space respectively, formulated as
\begin{flalign}
    v &= E_v(\Bar{v}) \\
    t &= E_t(\Bar{t_{ }})
\end{flalign}
Given paired training samples $\{v_i\}^N_{i=1}$ and $\{t_i\}^N_{i=1}$ where $v_i \in \mathcal{V}$ and  $t_i \in \mathcal{T}$, the video-text retrieval task can be defined as selecting the video $\Bar{v}_i$ which corresponds to the text query $\overline{t_{\text{query}}}$ between a given text query and a pool of candidate videos (Text-To-Video Retrieval, T2V), or vice versa (Video-To-Text Retrieval, V2T). Previous methods apply two projection functions $f_v$ and $f_t$ to project visual and textual space into a joint latent space and measure the distance within that, formulated as 
\begin{flalign}
   \text{T2V}(\overline{t_{\text{query}}}) =  \operatorname{argmin}_i\left(f_v(v_i), f_t(t_{query}) \right) \\
    \text{V2T}(\overline{v_{\text{query}}}) =  \operatorname{argmin}_j\left(f_v(v_{query})), f_t(t_j) \right) 
 \end{flalign}
where $t_{query} = D_t(\overline{t_{\text{query}}})$ and $v_{query} = D_v(\overline{v_{\text{query}}})$ are embeddings of the query samples.  

As depicted in Fig. \ref{LaT}, our goal is to learn the translation function between visual space $\mathcal{V}$ and textual space $\mathcal{T}$ under the guidance of learnt query embeddings $Q_G$ and $Q_F$. This leads to two translation relationships $G : t \longrightarrow \widetilde{v}$ and $F : v \longrightarrow \widetilde{t}$. Utilizing this latent translation function, we can enable cross-modal retrieval, formulated as:
\begin{flalign}
   \text{T2V}(\overline{t_{\text{query}}}) =  \operatorname{argmin}_i\left(v_i, G(t_{query}) \right) \\
    \text{V2T}(\overline{v_{\text{query}}}) =  \operatorname{argmin}_j\left(F(v_{query}), t_j \right) 
 \end{flalign}
In other words, we are able to translate textual embeddings from textual space $\mathcal{T}$ to visual space $\mathcal{V}$ and retrieve their most similar visual embedding from visual space $\mathcal{V}$, and vice versa.

\subsection{Model Architecture} \label{Model Architecture} 

Previous works in video-language modeling typically employ existing image and text encoders, such as the Clip-Based framework (CLIP4Clip~\cite{luo2021clip4clip},  CLIP2Video~\cite{fang2021clip2video} and CAMoE~\cite{cheng2021improving}), and the ViT-Bert-Based framework (HiT~\cite{liu2021hit}, Frozen~\cite{bain2021frozen}, and others~\cite{wang2021object}, ~\cite{yan2021video}). Due to the large pretraining dataset utilized by CLIP (400M), which would be difficult for us to replicate for our decoder's pretraining, we have chosen Frozen~\cite{bain2021frozen} (a ViT-Bert-Based framework) as our baseline\footnote{Another reason for adopting this structure is that the two encoders of CLIP are originally aligned, and its excellent zero-shot performance indicates that it can be easily adapted to the video-text dataset without any training. Furthermore, CLIP4CLIP shows that with CLIP weights, good video representations can be learned without temporal features (by using mean pooling). Therefore, we aim to reduce the modality gap in the alignment process. So we do not intend to use CLIP weights nor do we intend to compare with CLIP weight based methods.}. The visual encoder is an optimized TimeSformer~\cite{bertasius2021space} initialized with ViT~\cite{dosovitskiy2020image} weights trained on ImageNet-21k for spatial attention weights, and zero for temporal attention weights. The language encoder is DistilBERT base-uncased~\cite{sanh2019distilbert}.

As illustrated in Fig. \ref{LaT}, we have removed the projection layers after two encoders in order to eliminate any potential information distortion. In addition, we have added two decoders following DETR~\cite{carion2020end}'s approach to achieve cross-modal translation. The only difference between these two decoders is the number of queries stored in them, which has been adapted to the target modality. Additionally, due to GPU memory constraints, we have adopted the memory bank~\cite{chen2021empirical} to offset the decrease in batch size resulting from an increase of the network parameters.

\begin{figure}[t]
    \centering
    \includegraphics[width=.7\textwidth]{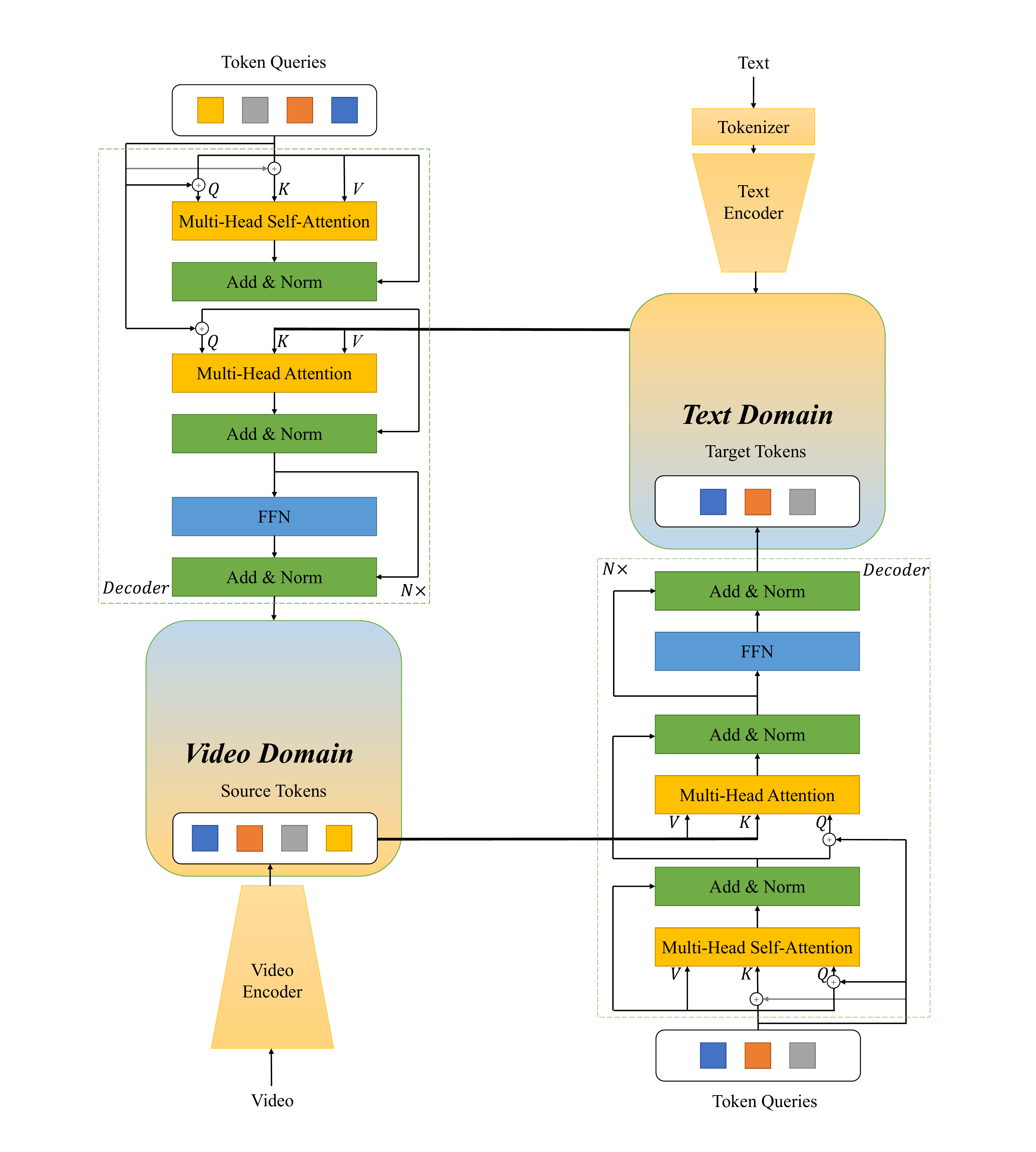}
    \caption{The detailed latent translation decoder. \textbf{Token queries} are some learnable parameters. The figure shows how source tokens in video domain translated to target tokens in text domain.}
    \label{detained_decoder}
\end{figure}

\subsection{Decoder} \label{decoder}

The detailed description of the decoder is given in Fig. \ref{detained_decoder}. This decoder follows the standard architecture of the DETR~\cite{carion2020end} transformer, transforming $M$ embeddings of size $d$ using multi-head self-attention and encoder-decoder attention mechanisms. Since the decoder is permutation-invariant, the $M$ input embeddings must be distinct to achieve different results. These input embeddings are learnt priors that we refer to as $token~ queries$, and we add them to the input of each attention layer. The perceiver resampler from Flamingo~\cite{alayrac2022flamingo} and Clipcap~\cite{mokady2021clipcap} also applied this framework.

Concretely, visual and textual features are extracted from the encoder and become the source tokens for multi-head attention, taking the roles of $V$ and $K$. For the decoder, there are learnable queries (the exact number depends on the target domain), which can initially be randomly initialized, passing through the multi-head self-attention before moving on to multi-head attention as $Q$. This process is repeated $N$ times and consequently yields target tokens that belong to the target domain. 

Intuitively, the queries act as multiple perspectives, transforming the source tokens to target tokens. The self-attention ensures that translations are distinct. We can implement additional constraints to ensure that the first query token captures global information (such as the global token from visual or textual embeddings if they exist, usually taken from the first or last token) while the other tokens capture more detailed information. This explains how our latent translation decoder works.

\subsection{Supervised Methods} \label{Loss}

Let $v \in \mathbb{R}^{N \times L_1 \times D}$ and $t \in \mathbb{R}^{N \times L_2 \times D}$ represent embeddings extracted from training samples in visual and textual modalities, where $L_1$ and $L_2$ denotes the number of tokens, $N$ denotes the batch size and $D$ denotes the dimension of each feature. $v \xrightarrow{F} t$ denote the translation network from visual modality to language modality, and $t \xrightarrow{G} v$ is the opposite. $v_{i,l_1}$ and $t_{i,l_2}$ denotes each embedding of each video-text pair, where $i = 1...N$, $l_1 = 1...L_1$, $l_2 = 1...L_2$ ($l_1 = 1$ or $l_2 = 1$ denotes the [CLS] token). 

\begin{equation*}
v = 
\begin{pmatrix}
v_{1,1} & v_{1,2} & \cdots & v_{1,L_1} \\
v_{2,1} & v_{2,2} & \cdots & v_{2,L_1} \\
\vdots  & \vdots  & \ddots & \vdots  \\
v_{N,1} & v_{N,2} & \cdots & v_{N,L_1} 
\end{pmatrix}
\end{equation*}

\begin{equation*}
t = 
\begin{pmatrix}
t_{1,1} & t_{1,2} & \cdots & t_{1,L_2} \\
t_{2,1} & t_{2,2} & \cdots & t_{2,L_2} \\
\vdots  & \vdots  & \ddots & \vdots  \\
t_{N,1} & t_{N,2} & \cdots & t_{N,L_2} 
\end{pmatrix}
\end{equation*}

First, we will provide an overview of our loss function and its components. Following this, we will explain how we address global and local information. Finally, we will present the overall training objective.

\subsubsection{Loss.}
Our loss includes inter-modal loss and intra-modal loss.

The inter-modal loss is utilized to learn the correct translation relationship. Follow the settings of Frozen~\cite{bain2021frozen}, we apply an infoNCE loss as the cross-modal loss.

\begin{equation}
    \mathcal{L}_{v2t}=-\frac{1}{N}\sum_{i=1}^{N} \text{log}\frac{\operatorname{exp}(sim(v_{i,l_1},t_{i,l_2})/\tau)}{\sum_{j=1}^{N}\operatorname{exp}(sim(v_{i,l_1},t_{j,l_2})/\tau)}
\end{equation}
\begin{equation}
    \mathcal{L}_{t2v}=-\frac{1}{N}\sum_{i=1}^{N} \text{log}\frac{\operatorname{exp}(sim(v_{i,l_1},t_{i,l_2})/\tau)}{\sum_{j=1}^{N}\operatorname{exp}(sim(v_{i,l_1},t_{j,l_2})/\tau)}
\end{equation}
where $\tau$ is the temperature and $sim$ is a similarity function(e.g., dot product). Specifically, when applying the general global infoNCE loss, $l_1=l_2=1$. And when applying the local infoNCE loss, there will be a MeanPooling method described in the next subsection.

As a result, the final inter-modal loss is 
\begin{equation}
    \mathcal{L}_{inter}=\frac{1}{2}\big(\mathcal{L}_{v2t}+\mathcal{L}_{t2v}\big)
\end{equation}

Intra-modal loss aims to make sure that after a cycle translation, the final target objects keep the same as the source target objects. To achieve this cycle-consistent propose, we apply a more strict constraint: mean square error (MSE) loss.

\begin{equation}
    \mathcal{L}_{intra}=\frac{1}{2}\big(||G(F(v_{i,l_1}))-v_{i,l_1}||_2^2+||F(G(t_{i,l_2}))-t_{i,l_2}||_2^2 \big )
\end{equation}
where $F$ denotes the translation network from visual modality to textual modality, and $G$ denotes the translation network form textual modality to visual modality.

\subsubsection{Global and Detailed Information} For paired videos and texts, passing them into their respective visual and textual encoders will obtain paired videos and texts embeddings, where there are multiple channels for each modality ($L > 1$). For instance, The output of visual transformer for an image input is usually set to $197 \times D$, with the first item of the 197 being the [CLS] token.

Previous cross-modal learning methods, such as CLIP~\cite{radford2021learning}, Frozen~\cite{bain2021frozen}, HiT~\cite{liu2021hit} tend to ignore the fine-grained information and solely interact through the global features ([CLS] token) of the entire video or sentence. We refer to this method, which only utilizes the [CLS] token, as global-level interaction.

\begin{equation}
    \mathcal{L}_{global}=\lambda_{inter}\mathcal{L}_{inter}+\lambda_{intra}\mathcal{L}_{intra}
\end{equation}
where $\lambda_{inter}$ and $\lambda_{intra}$ are two hyper-parameters to balance two objectives. We set both $\lambda_{inter}$ and $\lambda_{intra}$ to 1 in our experiments.

In addition, we are also concerned with the fine-grained information stored in other tokens. Recent works have proposed novel cross-modal interaction mechanisms to capture these fine-grained representations. For instance, FILIP~\cite{yao2021filip} uses a max-mean method for computing similarity, while~\cite{yan2021video} and~\cite{wang2021object} utilize auxiliary networks or existing object detection networks to assist similarity computation.

Here we just simply average all tokens except the [CLS] token and apply token-level loss as
\begin{equation}
    \mathcal{L}_{token}=\lambda_{inter}\mathcal{L}_{inter\_token}+\lambda_{intra}\mathcal{L}_{intra\_token}
\end{equation}

\subsubsection{Overall Training Objective.}
Thus, the overall objective function is the summary of global level loss and token level loss: 
\begin{equation}
    \mathcal{L}=\lambda_{global}\mathcal{L}_{global}+\lambda_{token}\mathcal{L}_{token}
\end{equation}
where $\lambda_{global}$ and $\lambda_{token}$ are two hyper-parameters to balance two objectives. We set both $\lambda_{global}$ and $\lambda_{token}$ to 1 in our experiments.

\section{Experiments}
In this section, we introduce the prertaining and finetuning datasets (Section~\ref{datasets}) and evaluation metrics (Section~\ref{metrics}) used in our experiments, as well as provide some implementation details (Section~\ref{implementation}). We then compare our performance with state-of-the-art methods to demonstrate the effectiveness and generality of our model (Section~\ref{comapre}) and conduct a series of ablation studies to understand the effects of the proposed components in our model (Section~\ref{ablation}). Finally, we provide further details on CLIP and LaT (Section~\ref{quan}) with visualizations for LaT (Section~\ref{visu}).

\subsection{Datasets} \label{datasets}
HowTo100M~\cite{miech2019howto100m} is a large-scale dataset of narrated videos with an emphasis on instructional videos where content creators teach complex tasks with an explicit intention of explaining the visual content on screen. It contains 136M video clips with captions sourced from 1.2M Youtube videos, and most existing works pretrain with it. However, HowTo100M is heavily noisy and only contains instructional videos, while WebVid2M~\cite{bain2021frozen} shows its efficiency with one tenth scale of the data. Meanwhile, although there is more than 10\% data from Google Conceptual Captions 3M (CC3M)~\cite{sharma2018conceptual} lost, applying the latest Google Conceptual Captions 12M (CC12M)~\cite{changpinyo2021conceptual} as our pretraining dataset is unfair for comparing with existing methods.

As a result, we conduct the experiments on 2 pretraining datasets (CC3M~\cite{sharma2018conceptual} and WebVid2M~\cite{bain2021frozen}) and 3 downstream datasets (MSR-VTT~\cite{xu2016msr}, 
MSVD~\cite{chen2011collecting}, 
DiDeMo~\cite{anne2017localizing}). The followings are the descriptions of these video-text (image-text) datasets.
\begin{itemize}
    \item[$\bullet$] \textbf{Conceptual Captions 3M (CC3M)~\cite{sharma2018conceptual}}  contains approximately 3.3 millions images annotated with captions. Conceptual Captions images and their raw descriptions are harvested from the web, and therefore represent a wider variety of styles. 
    \item[$\bullet$] \textbf{WebVid2M~\cite{bain2021frozen}} is a large scale text-video dataset, containing 2.5 millions video-text pairs scraped from the web. The videos are diverse and rich in their content.
    \item[$\bullet$] \textbf{MSR-VTT~\cite{xu2016msr}} contains 10,000 videos, where each video is annotated with 20 captions in English. We follow the training protocol defined in \cite{liu2021hit}\cite{bain2021frozen}\cite{luo2021clip4clip} to train on 9k videos, and to evaluate on text-to-video and video-to-text retrieval tasks on the 1k-A testing split with 1,000 video and text candidates defined by \cite{yu2018joint}.
    \item[$\bullet$] \textbf{MSVD~\cite{chen2011collecting}} contains 1,970 videos, and each video has approximately 40 captions in English and range form 1 to 62 seconds. The train, validation, and test splits contain 1,200, 100, and 670 videos, respectively.
    \item[$\bullet$] \textbf{DiDeMo~\cite{anne2017localizing}} contains 10k Flickr videos annotated with 40k sentences. We evaluate video-paragraph retrieval following \cite{lei2021less}, \cite{bain2021frozen}, \cite{luo2021clip4clip}, where all sentence descriptions for a video are concatenated into a single query.
\end{itemize}

\subsection{Evaluation Metrics} \label{metrics}
We use standard retrieval metrics: recall at rank K (R@K, K=1, 5, 10, higher is better), median rank (MedR, lower is better) to evaluate the performance of our model. 

\begin{table}[ht]
    \centering
    \caption{The experimental results on MSR-VTT.}
    \renewcommand\tabcolsep{3.0pt}
    \begin{tabular}{l |c c c c| c c c c}
        \toprule
          \multirow{2}{*}{\textbf{Methods}} & \multicolumn{4}{c|}{\textbf{Video-to-Text Retrieval}}& \multicolumn{4}{c}{\textbf{Text-to-Video Retrieval}} \\
        
        \cline{2-9} \\[-10pt]
        & \textbf{R@1} & \textbf{R@5} & \textbf{R@10} &\textbf{MedR} & \textbf{R@1} & \textbf{R@5} & \textbf{R@10} & \textbf{MedR} \\
        \midrule
         HowTo100M\cite{miech2019howto100m} & 12.2 & 33.5 & 47.5 &13 & 12.6 & 36.2 & 48.1 & 13 \\
         ActBert\cite{zhu2020actbert} & -- & -- & -- & -- & 16.3 &42.8 & 56.9 & 10 \\
         Noise-Estimation\cite{amrani2020noise} & -- & -- & -- & -- & 17.4 & 41.6 & 53.6 & 8 \\
         CE\cite{loshchilov2017decoupled} & 20.6 & 50.3 & 64.0 & 5.3 & 20.9 & 48.8 & 62.4 & 6 \\
         ClipBERT\cite{lei2021less} &--&--&--&--& 22.0&46.8&59.9& 6 \\
         MMT\cite{gabeur2020multi} &27.0 & 57.5 & 69.7 &3.7 &26.6 &57.1&69.6 &4 \\
         Support-Set\cite{patrick2020support} &28.5 &58.6 & 71.6 & 3 & 30.1 & 58.5 & 69.3 & 3 \\
         HiT\cite{liu2021hit} &32.1 & \textbf{62.7} & \textbf{74.1} &3 &30.7 &60.9&\textbf{73.2} & 3 \\
         Frozen\cite{bain2021frozen}  & -- & -- & -- & -- &31.0 & 59.5 & 70.5 & 3 \\ 
         \textbf{Ours(LaT)} & \textbf{35.4}& 61.3& 72.4& \textbf{3}&\textbf{35.3}& \textbf{61.3}& 72.9& \textbf{3}\\
        \midrule
         \multicolumn{1}{l}{\emph{Zero-shot}} \\
        \midrule
        Support-Set\cite{patrick2020support} & 8.7 & 23.0 & 31.1 & 31 & 12.7 & 27.5 & 36.2 & 24 \\
         Frozen\cite{bain2021frozen} & --&-- &-- &-- & 18.7 &39.5 & 51.6 &10 \\

         \textbf{Ours(LaT)} & \textbf{17.2}& \textbf{36.2}& \textbf{47.9}& \textbf{12}& \textbf{23.4}& \textbf{44.1}& \textbf{53.3}&\textbf{8} \\
         \bottomrule
    \end{tabular}
    
    \label{tab:msrvtt-compare}
\end{table}

\begin{table}[ht]
    \centering
    \renewcommand\tabcolsep{3.0pt}
    \caption{The experimental results on MSVD.}
    \begin{tabular}{l |c c c c| c c c c}
        \toprule
           \multirow{2}{*}{\textbf{Methods}} & \multicolumn{4}{c|}{\textbf{Video-to-Text Retrieval}}& \multicolumn{4}{c}{\textbf{Text-to-Video Retrieval}} \\
        \cline{2-9} \\[-10pt]
        & \textbf{R@1} & \textbf{R@5} & \textbf{R@10} &\textbf{MedR} & \textbf{R@1} & \textbf{R@5} & \textbf{R@10} & \textbf{MedR} \\
        \midrule
        CE\cite{loshchilov2017decoupled} & -- & -- & -- & --& 19.8 & 49.0 &63.8 & 6 \\
        Support-Set\cite{patrick2020support} & 34.7& 59.9& 70.0 & 3& 28.4 & 60.0 &72.9 &4 \\
        Frozen\cite{bain2021frozen}  & -- & -- & -- & --&33.7 & 64.7 & 76.3 & 3 \\ 
         \textbf{Ours(LaT)} &\textbf{39.7} &\textbf{75.6} &\textbf{85.4} &\textbf{2} &\textbf{40.0} &\textbf{74.6} &\textbf{84.2} &\textbf{2} \\
         \midrule
         \multicolumn{1}{l}{\emph{Zero-shot}} \\
         
         \midrule
        
         Frozen\cite{bain2021frozen}[Our Imp.]& 32.4 &65.5 &76.9 &3 &35.7 &63.9 &77.8 &3 \\
         
         \textbf{Ours(LaT)} &\textbf{34.4}& \textbf{69.0}& \textbf{79.2}&\textbf{3}& \textbf{36.9}&\textbf{68.6} &\textbf{81.0} &\textbf{2} \\
         \bottomrule
         
    \end{tabular}
    
    \label{tab:msvd-compare}
\end{table}

\subsection{Implementation Details} \label{implementation}
Following the code Bain~\etal~\cite{bain2021frozen} released, our experiments are conducted with PyTorch, optimized with Adam. We also apply the author's training strategy for pretraining, which means first training images and 1 frame videos with a learning rate of 3e-5, to capture the image content, then training 4 frames videos with a learning rate of 1e-5, to capture the video content. Experiments show that this schedule reduces both pretraining time and slightly improves pretraining results.

The whole pretraining takes 1 days on 8 Tesla V100 GPUs. Unless otherwise specified, all results reported in this paper regarding downstream datasets utilize the best pretraining model. Both during pretraining and fine-tuning, only 4 video frames are sampled.

\subsection{Compare to state of the art} \label{comapre}

\begin{table}[ht]
    \caption{The experimental results on DiDeMo.}
    \renewcommand\tabcolsep{3.0pt}
    \centering
    \begin{tabular}{l |c c c c| c c c c}
        \toprule
          \multirow{2}{*}{\textbf{Methods}} & \multicolumn{4}{c|}{\textbf{Video-to-Text Retrieval}}& \multicolumn{4}{c}{\textbf{Text-to-Video Retrieval}} \\
        \cline{2-9} \\[-10pt]
        & \textbf{R@1} & \textbf{R@5} & \textbf{R@10} &\textbf{MedR} & \textbf{R@1} & \textbf{R@5} & \textbf{R@10} & \textbf{MedR} \\
        \midrule
        ClipBERT\cite{lei2021less}& --&--&--&--&20.4 &44.5&56.7&7 \\
        Frozen\cite{bain2021frozen}  & -- & -- & -- & -- &31.0 & 59.8 & \textbf{72.4} & 3.0 \\ 
         \textbf{Ours(LaT)} &\textbf{32.7} &\textbf{61.1} &\textbf{72.7} &\textbf{3} &\textbf{32.6} &\textbf{61.3} &71.6 &\textbf{3} \\
         \midrule
         \multicolumn{1}{l}{\emph{Zero-shot}} \\
         
         \midrule
         Frozen\cite{bain2021frozen} & -- & --&-- & --& 21.1 &\textbf{46.0} & 56.2 &7 \\
         
         \textbf{Ours(LaT)} &\textbf{22.5} & \textbf{45.2}& \textbf{56.8}&\textbf{7} &\textbf{22.6} & 45.9&\textbf{58.9} &\textbf{7} \\
         \bottomrule
         
    \end{tabular}
    \label{tab:DiDeMo-compare}
\end{table}

We compare the proposed LaT with several vanilla state-of-the-art (SOTA) methods on MSR-VTT~\cite{xu2016msr}, MSVD~\cite{chen2011collecting} and DiDeMo~\cite{anne2017localizing} datasets and present the results in Table \ref{tab:msrvtt-compare}, \ref{tab:msvd-compare} and \ref{tab:DiDeMo-compare} respectively. 
The upper part of the tables display the fine-tuning results of methods. LaT is seen to outperform all comparison methods, especially in terms of R@1, in all three datasets. Furthermore, LaT showcases a minimal performance difference between video-to-text and text-to-video retrieval tasks (with a performance difference of R@1 of 0.1 by LaT versus 0.3 by CE and 0.4 by MMT in Table \ref{tab:msrvtt-compare}).

Apart from the fine-tuning results, we provide the results of zero-shot version of LaT. Compared with baseline methods Frozen~\cite{bain2021frozen} and Support-Set~\cite{patrick2020support}, the proposed LaT also shows better performance.

\subsection{Ablation Study} \label{ablation}

In this part, we study the effect of varying aspects of our latent translation framework to further demonstrate its effectiveness and robustness. All experiments are pretrained on WebVid2M with the same training epochs and batch size without any additional instruction. We report our text-to-video zero-shot result on MSR-VTT 1k-A testing split.

\noindent\textbf{Translation Methods} We investigate three translation methods, the \textbf{None} setting indicates no latent translation network is used, \textbf{Linear} is a 3-layer linear translation architecture, \textbf{Transformer} is a 3-layer transformer architecture, and \textbf{Decoder} is a 3-layer query-guided transformer (as shown in Figure~\ref{detained_decoder}) architecture. The results are given in Table \ref{tab:ablation-framework}, which reveal that the decoder framework produces better R@1 scores compared to other translation methods, and the incorporation of a simple transformer-based translation network delivers improved results compared to not having one.

\begin{table}[ht]
\caption{The ablation study on methods of latent translation.}
\renewcommand\tabcolsep{3.0pt}
    \centering
    \begin{tabular}{l c c c c}
        \toprule
        \textbf{Methods} & \textbf{R@1} & \textbf{R@5} & \textbf{R@10} & \textbf{MedR} \\
        \midrule
        None  & 19.4&38.6 &48.4 &11.5  \\
        
        Linear & 19.2&38.9 &46.9 &12 \\
        Transformer & 19.4& \textbf{41.2}& 50.4&10 \\
        Decoder & \textbf{20.3}& 41.1 &\textbf{51.4} &\textbf{9.5} \\
         \bottomrule
         
    \end{tabular}

    \label{tab:ablation-framework}
\end{table}

\noindent\textbf{Depth} We experiment with different depths of the translation framework by pre-training on WebVid2M for 200 epochs. The results are given in Table \ref{tab:ablation-depth}, which indicate an increase in recall rate as the number of layers increased. However, this increase in layers lead to an increase of model parameters and GFLOPs and a decrease in batch size. For example, when the depth of layers was 4, each GPU's batch size was decreased from 26 to 22 (compared to the depth being 3).

\begin{table}[ht]
\caption{The ablation study on depth of latent translation network. Depth = 4 utilizes a smaller batch size.}
\renewcommand\tabcolsep{3.0pt}
    \centering
    \begin{tabular}{c c c c c}
        \toprule
        \textbf{Depth} & \textbf{R@1} & \textbf{R@5} & \textbf{R@10} & \textbf{MedR} \\
        \midrule
        1  & 20.6 & 40.2& 50.0&10.5  \\
        
        2 & \textbf{20.7}& 40.7& 50.2&10 \\
        3 & 20.3& \textbf{41.1} &\textbf{51.4} &\textbf{9.5} \\
        4* & \textbf{20.7}& 39.9 & 48.2& 11 \\
         \bottomrule
         
    \end{tabular}
    \label{tab:ablation-depth}
\end{table}

\noindent\textbf{Number of Queries} We experiment with different number of queries of the translation framework by pretraining on WebVid2M with 100 training epochs. The results are given in Table \ref{tab:ablation-number_of_queries}, which show that when the number of guiding queries closer to the number of tokens in target domain, the better recall rate shows. In this experiment, the number of tokens in textual domain is close to 30 (The number of words from each text).

\begin{table}[ht]
\caption{The ablation study on number of queries. \bm{$N_{q}$} represents \textbf{Number of Queries}.}
\renewcommand\tabcolsep{3.0pt}
    \centering
    \begin{tabular}{c c c c c}
        \toprule
        \bm{$N_{q}$} & \textbf{R@1} & \textbf{R@5} & \textbf{R@10} & \textbf{MedR} \\
        \midrule
        15  & 19.6& 38.8& 47.5 &12  \\
        
        30 & \textbf{20.1}& \textbf{40.0} &\textbf{49.8} &\textbf{11} \\
        60 & 19.3& 37.7& 47.9& 12\\
         \bottomrule
         
    \end{tabular}
    
    \label{tab:ablation-number_of_queries}
\end{table}

\noindent\textbf{Token Usage} We experiment with different methods for utilizing detailed tokens by pretraining on WebVid2M with 100 training epochs. \textbf{Global} means using [CLS] token only. \textbf{Detailed} means to average all tokens except [CLS] token as the detailed token. The results are given in Table \ref{tab:ablation-tokne_usage}, which show that \textbf{Global} + \textbf{Detailed} works slightly better.

\begin{table}[ht]
\caption{The ablation study on token usage.}
\renewcommand\tabcolsep{3.0pt}
    \centering
    \begin{tabular}{c c c c c c}
        \toprule
        \textbf{Global} & \textbf{Detailed} & \textbf{R@1} & \textbf{R@5} & \textbf{R@10} & \textbf{MedR} \\
        \midrule
        \checkmark&  &20.1& \textbf{40.0} &\textbf{49.8} &\textbf{11}  \\
        
        &\checkmark & 20.3& 39.3& 49.3&\textbf{11} \\
        \checkmark &\checkmark & \textbf{20.5}& 39.8&49.7 &\textbf{11} \\
         \bottomrule

    \end{tabular}
    
    \label{tab:ablation-tokne_usage}
\end{table}

\noindent\textbf{Parameters Quantities} We experiment with different parameter quantites by pretraining on WebVid2M with 100 training epochs. The results are given in Table \ref{tab:ablation-para}, which show that our method does not rely on improvements in the number of parameters, but on better supervision.

\begin{table}[ht]
    \caption{The ablation study on parameters quantites.}
    \renewcommand\tabcolsep{3.0pt}
        \centering
        \begin{tabular}{c c c c c c}
            \toprule
            \textbf{Encoder Size} & \textbf{Decoder Size} & \textbf{R@1} & \textbf{R@5} & \textbf{R@10} & \textbf{MedR} \\
            \midrule
            6-layer-Bert& None &19.4& 38.6 &48.4 &11.5  \\
            12-layer-Bert & None & 19.9 & 40.8&50.4 & 10.0\\
            6-layer-Bert &2 3-layer-Decoder &\textbf{20.3}& \textbf{41.1} &\textbf{51.4} &\textbf{9.5} \\
             \bottomrule
    
        \end{tabular}
        \label{tab:ablation-para}
    \end{table}

\subsection{Quantities Details about CLIP and LaT}
\label{quan}
\begin{table}[ht]
    \centering
    \renewcommand\tabcolsep{1.0pt}
    \caption{Cosine Similarity in CLIP.}
    \begin{tabular}{c|c|ccc|ccc}
    \toprule
    \multicolumn{2}{c|}{\multirow{3}{*}{}} & \multicolumn{3}{c|}{\textbf{T}} & \multicolumn{3}{c}{\textbf{V}}  \\
        \cline{3-8} 
        \multicolumn{2}{c|}{}&\textbf{Apple}&\textbf{Dog}&\textbf{Man}&\textbf{Apple}&\textbf{Dog}&\textbf{Man}\\
         \midrule
         \multirow{3}{*}{\textbf{T}}&\textbf{Apple}&1.&0.84&0.77 &0.31 &0.21 &0.17  \\
         &\textbf{Dog} &0.84&1.&0.84&0.23 &0.29&0.17\\
         &\textbf{Man} &0.77&0.84&1.&0.21 &0.24&0.27\\
    \midrule
    \multirow{3}{*}{\textbf{V}}&\textbf{Apple}&0.31&0.23&0.21& 1.&0.72 &0.58 \\
         &\textbf{Dog} &0.21&0.29&0.24&0.72&1.&0.53\\
         &\textbf{Man}  &0.17&0.17&0.27&0.58&0.53&1.\\
    
    \bottomrule
    \end{tabular}
    \label{tab:2}
\end{table}

Table \ref{tab:2} shows \textbf{cosine similarity} (with larger values indicating greater similarity) between different embeddings in different latent space from CLIP. \textbf{T} denotes textual space and \textbf{V} denotes visual space. \textbf{Apple}, \textbf{Dog} and \textbf{Man} respectively denote embeddings with different semantic meanings from images or texts. The numerical values denote the cosine similarity between different embeddings. Additionally, it is worth noting that the cosine similarity for the same semantic embeddings between different latent spaces is significantly lower than those for different semantic embeddings within the same latent space, suggesting a substantial gap still exists between the visual and text space in CLIP. Liang~\etal~\cite{liang2022mind} also made similar observations.

An example that can be understood intuitively is: when two parallel planes intersect with a line perpendicular to them, two points arise, functioning as paired points of different modalities (planes). Through this approach, many paired points are obtained, each of which is the shortest distance away from the other. Nevertheless, this does not fuse the two modalities (planes) together.
\begin{table}[!ht]
    \centering
    \renewcommand\tabcolsep{1.0pt}
    \caption{Cosine Similarity in LaT.}
    \begin{tabular}{c|c|ccc|ccc|ccc|ccc}
    \toprule
    \multicolumn{2}{c|}{\multirow{3}{*}{}} & \multicolumn{3}{c|}{\textbf{T}} & \multicolumn{3}{c|}{\textbf{V}} & \multicolumn{3}{c|}{\textbf{GT}} & \multicolumn{3}{c}{\textbf{FV}} \\
        \cline{3-14} 
        \multicolumn{2}{c|}{}&\textbf{Apple}&\textbf{Dog}&\textbf{Man}&\textbf{Apple}&\textbf{Dog}&\textbf{Man}&\textbf{Apple}&\textbf{Dog}&\textbf{Man}&\textbf{Apple}&\textbf{Dog}&\textbf{Man} \\
         \midrule
         \multirow{3}{*}{\textbf{T}}&\textbf{Apple}&1.&0.67&0.53 &0.10 &0.02 &0.01 & 0.04&-0.00 &-0.02&0.59 &0.34 &0.38 \\
         &\textbf{Dog} &0.67&1.&0.59&0.08 &0.13&0.01&0.04&0.07&0.00&0.40 &0.59 &0.42\\
         &\textbf{Man} &0.53&0.59&1.&0.07 &0.01&0.07 &0.05&0.02&0.03&0.40 &0.30 &0.70\\
    \midrule
    \multirow{3}{*}{\textbf{V}}&\textbf{Apple}&0.10&0.08&0.07& 1.&0.47 &0.62 &0.57 &0.36&0.38 &0.15&0.05&0.09 \\
         &\textbf{Dog} &0.02&0.13&0.01&0.47&1.&0.47&0.29&0.60&0.27&0.03&0.17&0.02\\
         &\textbf{Man}  &0.01&0.01&0.07&0.62&0.47&1.&0.33&0.37&0.69&0.06&0.02&0.11\\
    \midrule
    \multirow{3}{*}{\textbf{GT}}&\textbf{Apple}&0.04&0.04&0.05& 0.57&0.29 &0.33 &1.&0.59&0.46 &0.03&0.06&0.02 \\
         &\textbf{Dog} &-0.00&0.07&0.02&0.36 &0.60 &0.37 &0.59&1.&0.55 &-0.01&0.06&0.02\\
         &\textbf{Man}  &-0.02&0.00&0.03& 0.38&0.27 &0.69 &0.46&0.55&1. &-0.05&0.01&0.03\\
    \midrule
    \multirow{3}{*}{\textbf{FV}}&\textbf{Apple}&0.59&0.40&0.40& 0.57&0.29&0.33 &0.03&-0.01&-0.05 &1.&0.47&0.62 \\
         &\textbf{Dog} &0.34&0.59&0.30& 0.36&0.60 &0.37 &0.06&0.06&0.01 &0.47&1.&0.47\\
         &\textbf{Man} &0.38&0.42&0.70& 0.38&0.27 &0.69 &0.02&0.02&0.03 &0.62&0.47&1.\\
    \bottomrule
    \end{tabular}
    \label{tab:1}
\end{table}

Table \ref{tab:1} shows \textbf{cosine similarity} (larger means more similar) between different embedding in different latent spaces from LaT. \textbf{T}, \textbf{V}, \textbf{GT}, and \textbf{FV} denote the textual space, visual space, visual space translated from textual space by the decoder $G$, and textual space translated from visual space by the decoder $F$, respectively. \textbf{Apple}, \textbf{Dog} and \textbf{Man} denote different embeddings with different semantic meanings from images or texts. The numbers denote the cosine similarity between different emebeddings. We can observe that, the cosine similarities for the same semantic embeddings between \textbf{GT} and \textbf{V} (or \textbf{FV} and \textbf{T}) are equal or higher than those of different semantic embeddings in the same latent space, which implies that LaT fuses different modalities better than the previous ways.

\subsection{Visualization on LaT}
\label{visu}
\begin{figure}[ht]
 
  \centering
\includegraphics[width=0.33\textwidth]{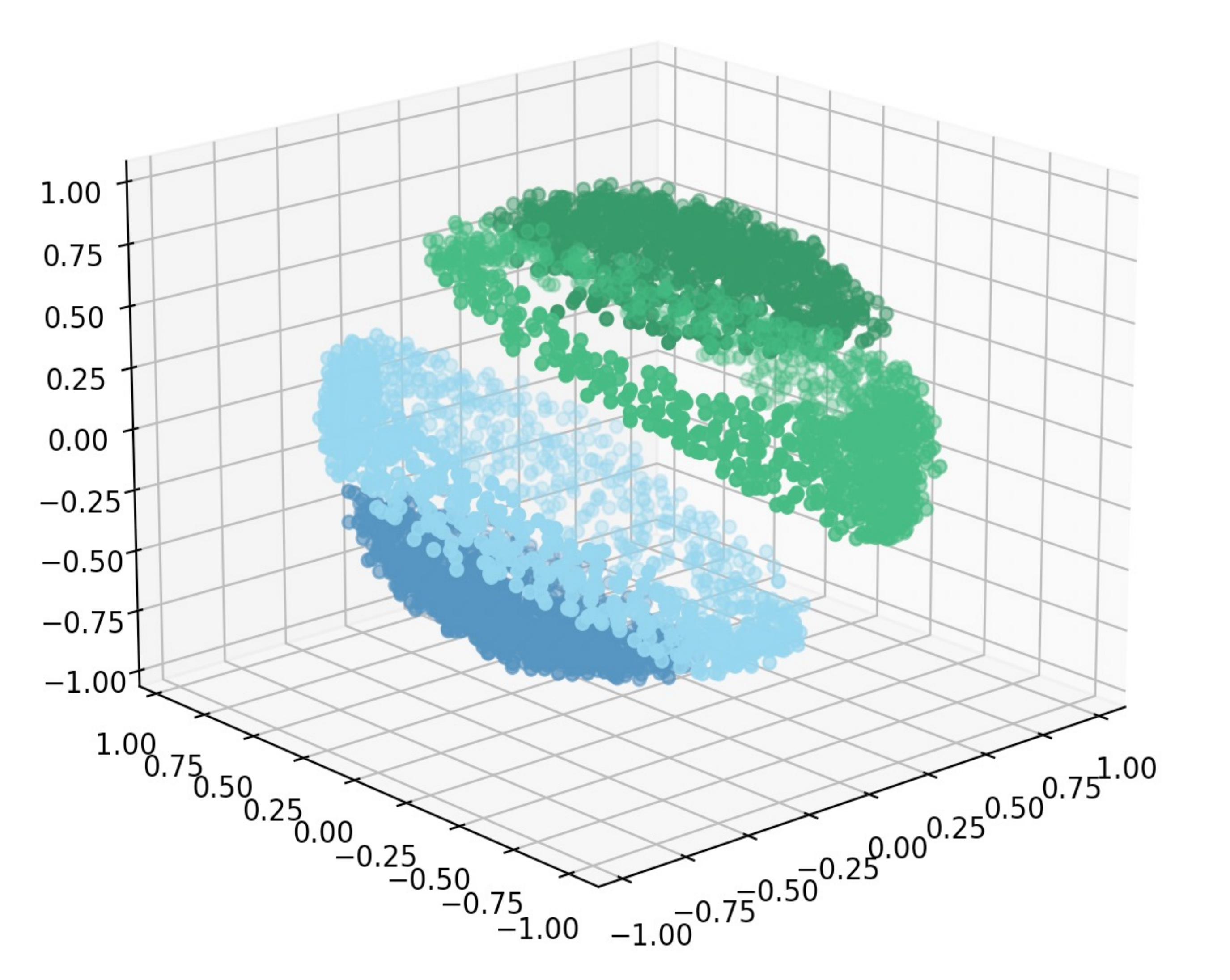}
  
  \caption{Visualization of embeddings from different latent space of LaT.} 
  \label{Visualization_LaT}
\end{figure}

Figure \ref{Visualization_LaT} demonstrates the visualization of embeddings from different latent space. The seagreen, steelblue, skyblue and mediumseagreen points represent latent embeddings from \textbf{T}, \textbf{V}, \textbf{GT} and \textbf{FV}, respectively. It can be clearly seen that LaT surpasses CLIP in merging \textbf{T} and \textbf{FV} (\textbf{V} and \textbf{GT}).

\section{Limitations and Discussions}

\begin{figure}
    \centering
    \includegraphics[width=.8\textwidth]{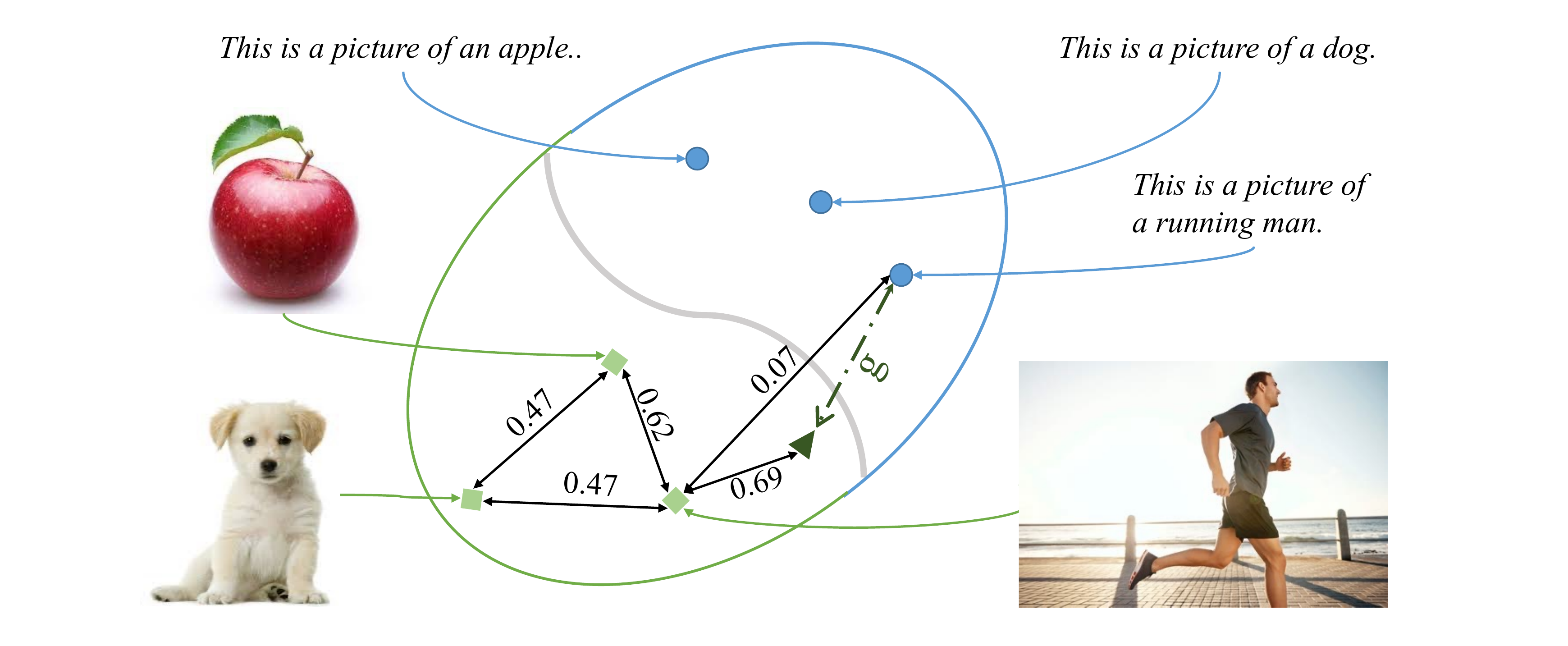}
    \caption{An example for showing the effectiveness of our LaT.}
    \label{LATVis}
\end{figure}

\noindent\textbf{Quantitative examples}: Figure~\ref{LATVis} provides more quantitative details about how our LaT works. The triangle represents a translation result from a text embedding through decoder $g$. Similar to figure-\ref{Visualization}-(b), we use \textbf{cosine similarity} (larger means more similar) as the measure. And it is evidently the picture \textit{Running man} is more similar to the translated text \textit{This is a picture of a running man} than picture \textit{Apple} and picture \textit{Dog}. To some extent, this means different modalities have been fused better than before.

\noindent\textbf{Fixed embedding space}: Although our method is successful in many cases, the results are far from perfect. During the training schedule, the parameters of both the encoders have been changed, so as latent space. In an ideal situation, these parameters would be fixed so that the latent translation network could be trained directly. We have explored this idea, but with little success thus far. Nonetheless, this will be the future work for us to explore a more suitable framework to achieve the latent translation, which is of great importance in large-scale cross-modal retrieval with existing encoders. Some existing works~\cite{gu2022wukong} have attempted to fix the visual encoder and achieve state-of-the-art performance on multiple downstream tasks. We believe, that with existing textual encoders and visual encoders, a learnable bridge can improve the efficiency of training and communication between different modalities. GLIDE~\cite{nichol2021glide} and DALL-E-2~\cite{ramesh2022hierarchical} from OpenAI have explored using the diffusion model~\cite{ho2020denoising,dhariwal2021diffusion} as the translation network between two latent spaces with success.

In addition, we found that freezing the backbones of both the visual encoder and textual encoder while retaining the learnability of tokenizer (which used to convert raw data to embedding input of backbone) will have a relatively comparable performance.

\noindent\textbf{Influence}: When doing search and recall tasks in the industry, the confidence of the recall results may be low. In this situation, it is hard to judge whether it is the modality gap of the model itself or the lack of high-quality recall data. Besides, reducing the modality gap can also bring more benefits to the modality-fused tasks, such as VQA and visual dialogues.

\section{Conclusions}

In this paper, we dive deep into the intrinsic gap between visual and textual domains with qualitative and quantitative analysis. And conjecture projecting them into a joint embedding space may lead to a distortion of intra-modal information. To tackle this challenges, we propose a latent translation mechanism (LaT) for cross-modal translation to solve the difficulty from aligning different modal spaces. Besides, we employ some learnable guiding parameters $Q$ for better translation. Extensive experiments conducted on MSR-VTT, MSVD, and DiDeMo datasets demonstrate the superiority and effectiveness of our LaT approach compared with vanilla state-of-the-art methods. Finally, we discuss the possible impacts and limitations of our approach.

\section{Acknowledgement}
We would like to thank Dr. Yi Zhang and anonymous friends for their kindly helpful discussions and feedback on writing.	

\end{sloppypar}

\bibliographystyle{splncs04}
\bibliography{egbib}

\begin{thebibliography}{10}
\providecommand{\url}[1]{\texttt{#1}}
\providecommand{\urlprefix}{URL }
\providecommand{\doi}[1]{https://doi.org/#1}

\bibitem{alayrac2022flamingo}
Alayrac, J.B., Donahue, J., Luc, P., Miech, A., Barr, I., Hasson, Y., Lenc, K.,
  Mensch, A., Millican, K., Reynolds, M., et~al.: Flamingo: a visual language
  model for few-shot learning. arXiv preprint arXiv:2204.14198  (2022)

\bibitem{amrani2020noise}
Amrani, E., Ben-Ari, R., Rotman, D., Bronstein, A.: Noise estimation using
  density estimation for self-supervised multimodal learning. arXiv preprint
  arXiv:2003.03186  \textbf{8} (2020)

\bibitem{anne2017localizing}
Anne~Hendricks, L., Wang, O., Shechtman, E., Sivic, J., Darrell, T., Russell,
  B.: Localizing moments in video with natural language. In: Proceedings of the
  IEEE international conference on computer vision. pp. 5803--5812 (2017)

\bibitem{arnab2021vivit}
Arnab, A., Dehghani, M., Heigold, G., Sun, C., Lu{\v{c}}i{\'c}, M., Schmid, C.:
  Vivit: A video vision transformer. In: Proceedings of the IEEE/CVF
  International Conference on Computer Vision. pp. 6836--6846 (2021)

\bibitem{bain2021frozen}
Bain, M., Nagrani, A., Varol, G., Zisserman, A.: Frozen in time: A joint video
  and image encoder for end-to-end retrieval. In: Proceedings of the IEEE/CVF
  International Conference on Computer Vision. pp. 1728--1738 (2021)

\bibitem{bertasius2021space}
Bertasius, G., Wang, H., Torresani, L.: Is space-time attention all you need
  for video understanding. arXiv preprint arXiv:2102.05095  \textbf{2}(3), ~4
  (2021)

\bibitem{borg2005modern}
Borg, I., Groenen, P.J.: Modern multidimensional scaling: Theory and
  applications. Springer Science \& Business Media (2005)

\bibitem{carion2020end}
Carion, N., Massa, F., Synnaeve, G., Usunier, N., Kirillov, A., Zagoruyko, S.:
  End-to-end object detection with transformers. In: European conference on
  computer vision. pp. 213--229. Springer (2020)

\bibitem{carreira2017quo}
Carreira, J., Zisserman, A.: Quo vadis, action recognition? a new model and the
  kinetics dataset. In: proceedings of the IEEE Conference on Computer Vision
  and Pattern Recognition. pp. 6299--6308 (2017)

\bibitem{changpinyo2021conceptual}
Changpinyo, S., Sharma, P., Ding, N., Soricut, R.: Conceptual 12m: Pushing
  web-scale image-text pre-training to recognize long-tail visual concepts. In:
  Proceedings of the IEEE/CVF Conference on Computer Vision and Pattern
  Recognition. pp. 3558--3568 (2021)

\bibitem{chen2011collecting}
Chen, D., Dolan, W.B.: Collecting highly parallel data for paraphrase
  evaluation. In: Proceedings of the 49th annual meeting of the association for
  computational linguistics: human language technologies. pp. 190--200 (2011)

\bibitem{chen2021empirical}
Chen, X., Xie, S., He, K.: An empirical study of training self-supervised
  vision transformers. In: Proceedings of the IEEE/CVF International Conference
  on Computer Vision. pp. 9640--9649 (2021)

\bibitem{chen2020uniter}
Chen, Y.C., Li, L., Yu, L., El~Kholy, A., Ahmed, F., Gan, Z., Cheng, Y., Liu,
  J.: Uniter: Universal image-text representation learning. In: European
  conference on computer vision. pp. 104--120. Springer (2020)

\bibitem{cheng2021improving}
Cheng, X., Lin, H., Wu, X., Yang, F., Shen, D.: Improving video-text retrieval
  by multi-stream corpus alignment and dual softmax loss. arXiv preprint
  arXiv:2109.04290  (2021)

\bibitem{cornia2018towards}
Cornia, M., Baraldi, L., Tavakoli, H.R., Cucchiara, R.: Towards
  cycle-consistent models for text and image retrieval. In: Proceedings of the
  European Conference on Computer Vision (ECCV) Workshops. pp.~0--0 (2018)

\bibitem{dhariwal2021diffusion}
Dhariwal, P., Nichol, A.: Diffusion models beat gans on image synthesis.
  Advances in Neural Information Processing Systems  \textbf{34},  8780--8794
  (2021)

\bibitem{dosovitskiy2020image}
Dosovitskiy, A., Beyer, L., Kolesnikov, A., Weissenborn, D., Zhai, X.,
  Unterthiner, T., Dehghani, M., Minderer, M., Heigold, G., Gelly, S., et~al.:
  An image is worth 16x16 words: Transformers for image recognition at scale.
  arXiv preprint arXiv:2010.11929  (2020)

\bibitem{fang2021clip2video}
Fang, H., Xiong, P., Xu, L., Chen, Y.: Clip2video: Mastering video-text
  retrieval via image clip. arXiv preprint arXiv:2106.11097  (2021)

\bibitem{gabeur2020multi}
Gabeur, V., Sun, C., Alahari, K., Schmid, C.: Multi-modal transformer for video
  retrieval. In: European Conference on Computer Vision. pp. 214--229. Springer
  (2020)

\bibitem{goodfellow2014generative}
Goodfellow, I., Pouget-Abadie, J., Mirza, M., Xu, B., Warde-Farley, D., Ozair,
  S., Courville, A., Bengio, Y.: Generative adversarial nets. Advances in
  neural information processing systems  \textbf{27} (2014)

\bibitem{gu2022wukong}
Gu, J., Meng, X., Lu, G., Hou, L., Niu, M., Xu, H., Liang, X., Zhang, W.,
  Jiang, X., Xu, C.: Wukong: 100 million large-scale chinese cross-modal
  pre-training dataset and a foundation framework. arXiv preprint
  arXiv:2202.06767  (2022)

\bibitem{ho2020denoising}
Ho, J., Jain, A., Abbeel, P.: Denoising diffusion probabilistic models.
  Advances in Neural Information Processing Systems  \textbf{33},  6840--6851
  (2020)

\bibitem{hu2021unit}
Hu, R., Singh, A.: Unit: Multimodal multitask learning with a unified
  transformer. In: Proceedings of the IEEE/CVF International Conference on
  Computer Vision. pp. 1439--1449 (2021)

\bibitem{ji20123d}
Ji, S., Xu, W., Yang, M., Yu, K.: 3d convolutional neural networks for human
  action recognition. IEEE transactions on pattern analysis and machine
  intelligence  \textbf{35}(1),  221--231 (2012)

\bibitem{jia2021scaling}
Jia, C., Yang, Y., Xia, Y., Chen, Y.T., Parekh, Z., Pham, H., Le, Q., Sung,
  Y.H., Li, Z., Duerig, T.: Scaling up visual and vision-language
  representation learning with noisy text supervision. In: International
  Conference on Machine Learning. pp. 4904--4916. PMLR (2021)

\bibitem{karpathy2014large}
Karpathy, A., Toderici, G., Shetty, S., Leung, T., Sukthankar, R., Fei-Fei, L.:
  Large-scale video classification with convolutional neural networks. In:
  Proceedings of the IEEE conference on Computer Vision and Pattern
  Recognition. pp. 1725--1732 (2014)

\bibitem{lei2021less}
Lei, J., Li, L., Zhou, L., Gan, Z., Berg, T.L., Bansal, M., Liu, J.: Less is
  more: Clipbert for video-and-language learning via sparse sampling. In:
  Proceedings of the IEEE/CVF Conference on Computer Vision and Pattern
  Recognition. pp. 7331--7341 (2021)

\bibitem{li2020unicoder}
Li, G., Duan, N., Fang, Y., Gong, M., Jiang, D.: Unicoder-vl: A universal
  encoder for vision and language by cross-modal pre-training. In: Proceedings
  of the AAAI Conference on Artificial Intelligence. vol.~34, pp. 11336--11344
  (2020)

\bibitem{li2019visualbert}
Li, L.H., Yatskar, M., Yin, D., Hsieh, C.J., Chang, K.W.: Visualbert: A simple
  and performant baseline for vision and language. arXiv preprint
  arXiv:1908.03557  (2019)

\bibitem{li2020oscar}
Li, X., Yin, X., Li, C., Zhang, P., Hu, X., Zhang, L., Wang, L., Hu, H., Dong,
  L., Wei, F., et~al.: Oscar: Object-semantics aligned pre-training for
  vision-language tasks. In: European Conference on Computer Vision. pp.
  121--137. Springer (2020)

\bibitem{liang2022mind}
Liang, W., Zhang, Y., Kwon, Y., Yeung, S., Zou, J.: Mind the gap: Understanding
  the modality gap in multi-modal contrastive representation learning. arXiv
  preprint arXiv:2203.02053  (2022)

\bibitem{lin2021vx2text}
Lin, X., Bertasius, G., Wang, J., Chang, S.F., Parikh, D., Torresani, L.:
  Vx2text: End-to-end learning of video-based text generation from multimodal
  inputs. In: Proceedings of the IEEE/CVF Conference on Computer Vision and
  Pattern Recognition. pp. 7005--7015 (2021)

\bibitem{liu2021selective}
Liu, C., Li, X., Chen, H., Modolo, D., Tighe, J.: Selective feature compression
  for efficient activity recognition inference. In: Proceedings of the IEEE/CVF
  International Conference on Computer Vision. pp. 13628--13637 (2021)

\bibitem{liu2021hit}
Liu, S., Fan, H., Qian, S., Chen, Y., Ding, W., Wang, Z.: Hit: Hierarchical
  transformer with momentum contrast for video-text retrieval. In: Proceedings
  of the IEEE/CVF International Conference on Computer Vision. pp. 11915--11925
  (2021)

\bibitem{liu2019cyclematch}
Liu, Y., Guo, Y., Liu, L., Bakker, E.M., Lew, M.S.: Cyclematch: A
  cycle-consistent embedding network for image-text matching. Pattern
  Recognition  \textbf{93},  365--379 (2019)

\bibitem{loshchilov2017decoupled}
Loshchilov, I., Hutter, F.: Decoupled weight decay regularization. arXiv
  preprint arXiv:1711.05101  (2017)

\bibitem{lu2019vilbert}
Lu, J., Batra, D., Parikh, D., Lee, S.: Vilbert: Pretraining task-agnostic
  visiolinguistic representations for vision-and-language tasks. Advances in
  neural information processing systems  \textbf{32} (2019)

\bibitem{luo2021clip4clip}
Luo, H., Ji, L., Zhong, M., Chen, Y., Lei, W., Duan, N., Li, T.: Clip4clip: An
  empirical study of clip for end to end video clip retrieval. arXiv preprint
  arXiv:2104.08860  (2021)

\bibitem{miech2019howto100m}
Miech, A., Zhukov, D., Alayrac, J.B., Tapaswi, M., Laptev, I., Sivic, J.:
  Howto100m: Learning a text-video embedding by watching hundred million
  narrated video clips. In: Proceedings of the IEEE/CVF International
  Conference on Computer Vision. pp. 2630--2640 (2019)

\bibitem{mokady2021clipcap}
Mokady, R., Hertz, A., Bermano, A.H.: Clipcap: Clip prefix for image
  captioning. arXiv preprint arXiv:2111.09734  (2021)

\bibitem{nichol2021glide}
Nichol, A., Dhariwal, P., Ramesh, A., Shyam, P., Mishkin, P., McGrew, B.,
  Sutskever, I., Chen, M.: Glide: Towards photorealistic image generation and
  editing with text-guided diffusion models. arXiv preprint arXiv:2112.10741
  (2021)

\bibitem{patrick2020support}
Patrick, M., Huang, P.Y., Asano, Y., Metze, F., Hauptmann, A., Henriques, J.,
  Vedaldi, A.: Support-set bottlenecks for video-text representation learning.
  arXiv preprint arXiv:2010.02824  (2020)

\bibitem{radford2021learning}
Radford, A., Kim, J.W., Hallacy, C., Ramesh, A., Goh, G., Agarwal, S., Sastry,
  G., Askell, A., Mishkin, P., Clark, J., et~al.: Learning transferable visual
  models from natural language supervision. In: International Conference on
  Machine Learning. pp. 8748--8763. PMLR (2021)

\bibitem{ramesh2022hierarchical}
Ramesh, A., Dhariwal, P., Nichol, A., Chu, C., Chen, M.: Hierarchical
  text-conditional image generation with clip latents. arXiv preprint
  arXiv:2204.06125  (2022)

\bibitem{sanh2019distilbert}
Sanh, V., Debut, L., Chaumond, J., Wolf, T.: Distilbert, a distilled version of
  bert: smaller, faster, cheaper and lighter. arXiv preprint arXiv:1910.01108
  (2019)

\bibitem{sharma2018conceptual}
Sharma, P., Ding, N., Goodman, S., Soricut, R.: Conceptual captions: A cleaned,
  hypernymed, image alt-text dataset for automatic image captioning. In:
  Proceedings of the 56th Annual Meeting of the Association for Computational
  Linguistics (Volume 1: Long Papers). pp. 2556--2565 (2018)

\bibitem{sun2019videobert}
Sun, C., Myers, A., Vondrick, C., Murphy, K., Schmid, C.: Videobert: A joint
  model for video and language representation learning. In: Proceedings of the
  IEEE/CVF International Conference on Computer Vision. pp. 7464--7473 (2019)

\bibitem{tran2015learning}
Tran, D., Bourdev, L., Fergus, R., Torresani, L., Paluri, M.: Learning
  spatiotemporal features with 3d convolutional networks. In: Proceedings of
  the IEEE international conference on computer vision. pp. 4489--4497 (2015)

\bibitem{tran2018closer}
Tran, D., Wang, H., Torresani, L., Ray, J., LeCun, Y., Paluri, M.: A closer
  look at spatiotemporal convolutions for action recognition. In: Proceedings
  of the IEEE conference on Computer Vision and Pattern Recognition. pp.
  6450--6459 (2018)

\bibitem{vaswani2017attention}
Vaswani, A., Shazeer, N., Parmar, N., Uszkoreit, J., Jones, L., Gomez, A.N.,
  Kaiser, {\L}., Polosukhin, I.: Attention is all you need. Advances in neural
  information processing systems  (2017)

\bibitem{wang2021object}
Wang, A.J., Ge, Y., Cai, G., Yan, R., Lin, X., Shan, Y., Qie, X., Shou, M.Z.:
  Object-aware video-language pre-training for retrieval. arXiv preprint
  arXiv:2112.00656  (2021)

\bibitem{wang2021efficientclip}
Wang, J., Wang, H., Deng, J., Wu, W., Zhang, D.: Efficientclip: Efficient
  cross-modal pre-training by ensemble confident learning and language
  modeling. arXiv preprint arXiv:2109.04699  (2021)

\bibitem{wang2016temporal}
Wang, L., Xiong, Y., Wang, Z., Qiao, Y., Lin, D., Tang, X., Gool, L.V.:
  Temporal segment networks: Towards good practices for deep action
  recognition. In: European conference on computer vision. pp. 20--36. Springer
  (2016)

\bibitem{wu2018cycle}
Wu, L., Wang, Y., Shao, L.: Cycle-consistent deep generative hashing for
  cross-modal retrieval. IEEE Transactions on Image Processing  \textbf{28}(4),
   1602--1612 (2018)

\bibitem{xu2016msr}
Xu, J., Mei, T., Yao, T., Rui, Y.: Msr-vtt: A large video description dataset
  for bridging video and language. In: Proceedings of the IEEE conference on
  computer vision and pattern recognition. pp. 5288--5296 (2016)

\bibitem{yan2021video}
Yan, R., Shou, M.Z., Ge, Y., Wang, A.J., Lin, X., Cai, G., Tang, J.: Video-text
  pre-training with learned regions. arXiv preprint arXiv:2112.01194  (2021)

\bibitem{yao2021filip}
Yao, L., Huang, R., Hou, L., Lu, G., Niu, M., Xu, H., Liang, X., Li, Z., Jiang,
  X., Xu, C.: Filip: Fine-grained interactive language-image pre-training.
  arXiv preprint arXiv:2111.07783  (2021)

\bibitem{yu2018joint}
Yu, Y., Kim, J., Kim, G.: A joint sequence fusion model for video question
  answering and retrieval. In: Proceedings of the European Conference on
  Computer Vision (ECCV). pp. 471--487 (2018)

\bibitem{zhu2017unpaired}
Zhu, J.Y., Park, T., Isola, P., Efros, A.A.: Unpaired image-to-image
  translation using cycle-consistent adversarial networks. In: Proceedings of
  the IEEE international conference on computer vision. pp. 2223--2232 (2017)

\bibitem{zhu2020actbert}
Zhu, L., Yang, Y.: Actbert: Learning global-local video-text representations.
  In: Proceedings of the IEEE/CVF conference on computer vision and pattern
  recognition. pp. 8746--8755 (2020)

\end{thebibliography}
\end{document}